\documentclass[sigconf]{acmart}
\AtBeginDocument{%
  }

\setcopyright{acmlicensed}
\copyrightyear{2018}
\acmYear{2018}
\acmDOI{XXXXXXX.XXXXXXX}
\acmConference[Conference acronym 'XX]{Make sure to enter the correct
  conference title from your rights confirmation email}{June 03--05,
  2018}{Woodstock, NY}
\acmISBN{978-1-4503-XXXX-X/2018/06}

\usepackage{multirow}    
\usepackage{xspace}
\usepackage{mathtools}
\usepackage{pifont}
\usepackage[table]{xcolor}

\begin{document}


\title{Adaptive Reciprocal Knowledge Distillation}
\title{Discovering and Preserving Category Correlation Knowledge via Adaptive Reciprocal Knowledge Distillation}
\author{Dawen Jiang}
\affiliation{%
  \institution{Wuhan University of Technology}
  \city{Wuhan}
  \state{Hubei} 
  \country{China}}
\email{davin@whut.edu.cn}
\orcid{0009-0006-5898-4403}

\author{Zhishu Shen}
\authornote{Zhishu Shen is the corresponding author.}
\affiliation{%
  \institution{Wuhan University of Technology}
  \city{Wuhan}
  \state{Hubei}
  \country{China}}
\email{z\_shen@ieee.org}
\orcid{0000-0002-3123-4390}

\author{Zeyu Liu}
\affiliation{%
  \institution{Wuhan University of Technology}
  \city{Wuhan}
  \state{Hubei}
  \country{China}
}
 \email{357932@whut.edu.cn}
 \orcid{0009-0009-5083-3856}

\author{Tiehua Zhang}
\affiliation{%
 \institution{Tongji University}
 \city{Shanghai}
 \country{China}}
 \email{tiehuaz@tongji.edu.cn}
\orcid{0000-0002-7195-4472}

\newcommand{\algname}{AR-KD\xspace}
\newcommand*{\note}[1]{\textcolor{red}{#1}}
\newcommand*{\shen}[1]{\textcolor{violet}{#1}}
\newcommand*{\jdw}[1]{\textcolor{blue}{#1}}
\renewcommand{\shortauthors}{Dawen Jiang, Zhishu Shen, Zeyu Liu and Tiehua Zhang}

\renewcommand{\textcolor}[2]{#2}
\begin{abstract}
Knowledge distillation aims to improve the performance of lightweight student models by transferring knowledge from larger and more powerful teacher models. However, a substantial size gap between teacher and student models often impedes effective knowledge transfer. Most existing approaches adopt a static, one-way teacher-to-student distillation paradigm, which overlooks the dynamic nature of student learning and fails to provide targeted guidance on hard samples. In this paper, we propose adaptive reciprocal knowledge distillation (\algname), a novel method that improves knowledge transfer by simplifying the teacher’s output distribution. \jdw{Specifically, \algname performs reciprocal adaptation on the teacher by matching its class correlation matrix to the student’s relational representation, which reshapes the teacher’s prediction structure to better suit the student’s capacity. This relational alignment mitigates the collapse of inter-class dark knowledge caused by overconfident teachers, enabling the student to learn from richer and more compatible supervisory signals. We evaluate \algname on CIFAR-100 and ImageNet-1k classification datasets, where it outperforms state-of-the-art knowledge distillation baselines. Specifically, \algname improves student performance across homogeneous and heterogeneous setups: up to 7.13\% accuracy gain for students, 1.42\%–4.15\% higher than vanilla KD on average, and further improvements when integrated with other advanced methods.}  Our code is available at https://anonymous.4open.science/r/ARKD/.
\end{abstract}




\maketitle

\section{Introduction}

Deep neural networks (DNNs)~\cite{resnet_he,imagenet2012,VGG} have achieved remarkable success in various computer vision tasks. Models such as ResNet and Vision Transformers have significantly pushed the boundaries of accuracy by leveraging large-scale architectures and massive data. However, the impressive performance of these models comes at the cost of enormous computational and memory demands. To tackle this challenge, a variety of model compression methods have been proposed. Knowledge distillation (KD)~\cite{KD_Hinton} has emerged as one of the most effective and generalizable strategies. KD aims to transfer the knowledge from a large, well-trained teacher model to a smaller student model by minimizing the Kullback-Leibler (KL) divergence between their outputs. To alleviate distillation degradation caused by large teacher--student capacity gaps, Pei \textit{et al.} proposed self-boosting feature distillation (SFD)~\cite{pei2021selfboosting}, which enhances student representations through feature integration and parameter self-distillation without additional overhead. Similarly, Kim \textit{et al.} introduced integrating matched features (IMF) using attentive logit~\cite{kim2023imf}, leveraging intermediate feature distillers and an ENA layer to effectively integrate matched features, achieving improved performance across diverse tasks. 

Numerous KD methods have been developed to improve the performance of compact student models by leveraging the guidance of pre-trained teachers. These methods generally fall into two categories: logits-based distillation~\cite{KD_Hinton,TAKD,DGKD,DKD_2022,ResKD,Muti_teacher_KD,MTKD_RL,MLKD,Logit_CTKD,RCKA,logit_LSKD}, which transfers softened output probabilities from teacher to student, and feature-based distillation~\cite{feature_AT,feature_CRD,feature_FitNet,feature_OFD,feature_relationKD,feature_ReviewKD,feature_catkd,feature_kr}, which aligns intermediate representations between models. These methods have consistently demonstrated superior performance compared to training student models from scratch. \note{However, a substantial architectural gap often hinders effective knowledge transfer, as the student struggles to mimic the teacher’s overconfident and complex output distributions.}

To bridge this gap, recent works have proposed approaches such as softening the teacher outputs, introducing assistant teacher models, and employing multi-teacher distillation. However, most existing methods adopt a static, one-way teacher-to-student distillation paradigm, which overlooks the dynamic nature of the student’s learning process. \note{Moreover, they focus more on enabling the student to capture the teacher’s correctly predicted classes, while ignoring the correlation between categories. This issue often prevents the student from correctly capturing category correlations. In large-scale classification tasks such as e-commerce product categorization, document topic annotation, and medical code prediction, the category correlation information is essential.}

\begin{figure*}[tb!]
    \centering
    \includegraphics[width=0.65\linewidth]{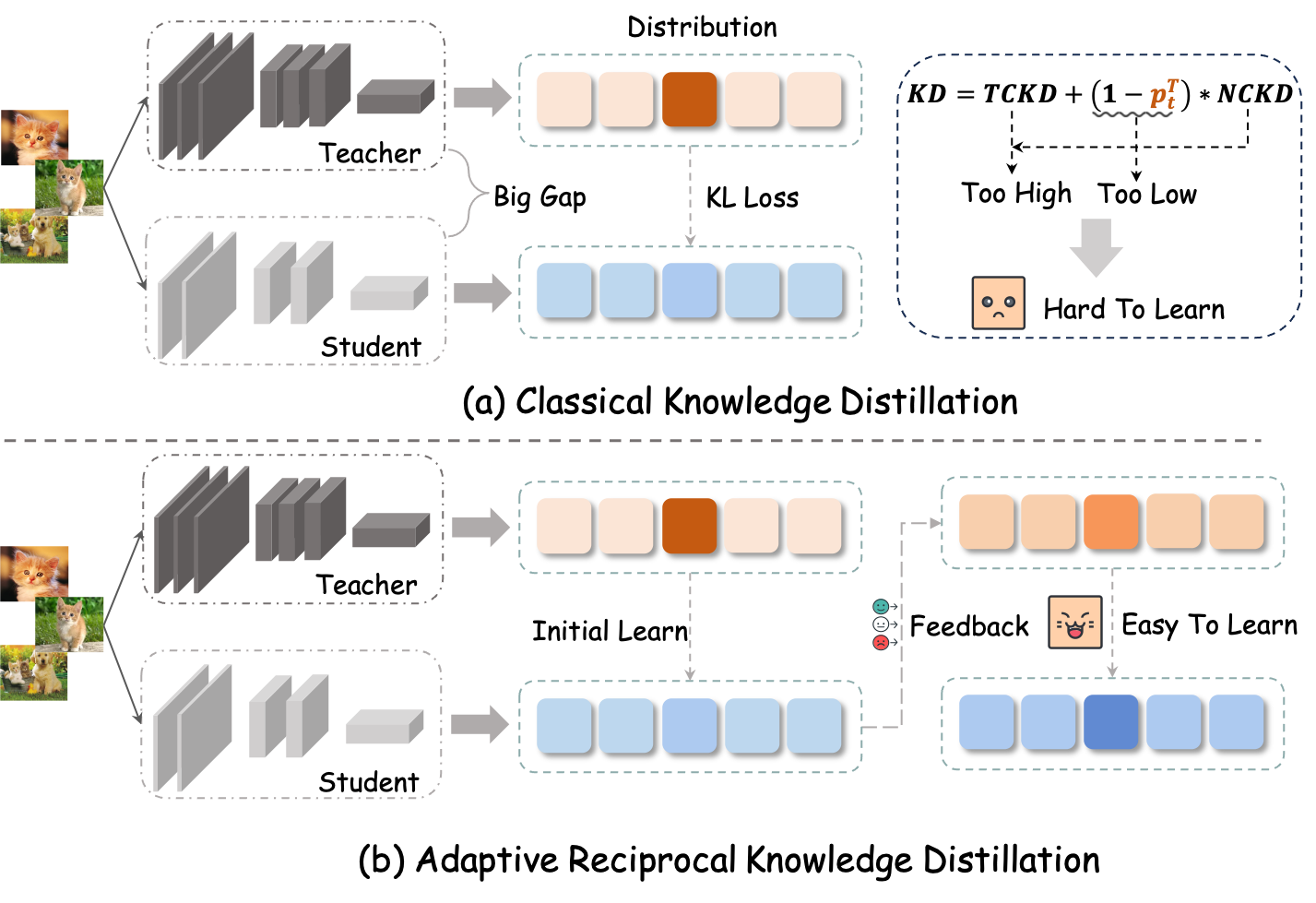}
    \caption{The core motivation of our \algname lies in addressing the capacity gap in traditional KD. According to \protect\cite{DKD_2022}, the high confidence of teacher outputs results in low weights for non-target class knowledge (NCKD), making it difficult for the student to effectively learn from the dark knowledge. In contrast, \algname introduces a reciprocal distillation strategy that adaptively simplifies the teacher outputs based on feedback from the student, reducing overconfidence and softening the knowledge, making knowledge transfer more effective and student-friendly.}
    \label{fig:figure1}
\end{figure*}

In this paper, we propose adaptive reciprocal knowledge distillation (\algname), a novel logits-based distillation framework that enhances the compatibility between teacher and student through reciprocal model adaptation. As illustrated in \figurename~\ref{fig:figure1}, instead of following a conventional one-way teacher-to-student paradigm, \algname introduces a bidirectional training scheme in which the student also provides feedback to guide the teacher, forming a closed-loop knowledge refinement process. Specifically, \algname adopts a recursive distillation pipeline that integrates a reciprocal teacher adaptation mechanism into standard KD. In the first stage, the student is trained with the original teacher to obtain an initial representation of the task. In the second stage, the teacher is further adapted by aligning its category correlation structure with that of the student. This is achieved by matching the class correlation matrices derived from their softened prediction distributions, which allows the teacher to adjust its relational knowledge to better fit the student’s capacity. Through this reciprocal adaptation, the teacher learns to generate more compatible and transferable outputs for the student.
In the final stage, the adapted teacher is used to distill knowledge back to the student, enabling the student to learn from a teacher whose category correlations are explicitly aligned with its own representation. This bidirectional relationally-aware design allows the student to indirectly influence the teacher and leads to a more adaptive and effective distillation process. Moreover, since \algname operates at the level of class correlation structure, it can be seamlessly integrated with existing KD methods as a plug-and-play enhancement module.

The main contributions of this paper are summarized as follows:
\begin{itemize}

    \item We propose a novel logits-based distillation framework \algname that transforms knowledge transfer into a closed-loop process. Rather than treating the teacher as a static supervisor, our approach explicitly allows the student to adapt the teacher, thereby discovering and preserving essential category correlation knowledge that would otherwise be obscured by teacher overconfidence.

    \item \note{We design a relational structure alignment strategy that matches the category correlation matrices derived from the softened prediction distributions of both teacher and student models. This mechanism enables the teacher to dynamically adjust its relational knowledge, generating highly compatible and transferable outputs tailored to the student's representation capability.}

    \item We validate the effectiveness of our method through extensive experiments conducted on CIFAR-100 and ImageNet-1k against state-of-the-art baselines under both homogeneous and heterogeneous teacher–student configurations.
\end{itemize}

\section{Related Work}
KD aims to transfer knowledge from a large, complex teacher model to a lightweight student model, improving the student's performance with fewer parameters and reduced computational cost~\cite{LiTNNLS25,TianCSUR25,JiangIoTM25}. Since its introduction, numerous studies have been conducted based on this concept. Current distillation methods can be broadly divided into two categories: logit-based distillation~\cite{KD_Hinton,TAKD,DGKD,DKD_2022,ResKD,Muti_teacher_KD,MTKD_RL,MLKD,Logit_CTKD,logit_LSKD} and intermediate feature-based distillation~\cite{feature_AT,feature_CRD,feature_FitNet,feature_OFD,feature_relationKD,feature_ReviewKD,feature_catkd,feature_kr}. 
\textcolor{blue}{Sun \textit{et al.}~\cite{SUN2025111095} proposed Exp-KD, which adopts class activation map (CAM) as the knowledge carrier to unify label-dependent and structure-related knowledge, thereby improving interpretability. Experiments on CIFAR-100 and ImageNet-1K demonstrate its superior performance.} 
\textcolor{blue}{Gou \textit{et al.}~\cite{Gou2024HMAT} proposed a  hierarchical multi-attention transfer framework (HMAT) to transfer hierarchical multi-attention knowledge in feature-based distillation, achieving state-of-the-art performance on multiple vision tasks.} 
Most KD methods rely on a pre-trained, complex teacher to supervise a simpler student. However, when the capacity gap is excessively large, the student may struggle to absorb the distilled knowledge, limiting distillation effectiveness.

To address this issue, several knowledge distillation methods have been proposed. For example, teacher assistant knowledge distillation (TAKD)~\cite{TAKD} introduces an intermediate teacher assistant to reduce the capacity gap, but may propagate teacher errors multiple times, causing error accumulation. To alleviate this problem, densely guided knowledge distillation (DGKD)~\cite{DGKD} employs tightly guided supervision to reduce compounded errors. Decoupled knowledge distillation (DKD)~\cite{DKD_2022} tackles performance degradation by decoupling the distillation loss and introducing hyperparameters to better handle large teacher--student gaps. 
\textcolor{blue}{Similarly, Yang \textit{et al.}~\cite{Yang_2023_ICCV} proposed a unified KD and self-KD framework with NKD and USKD, compatible with both CNN and ViT architectures and achieving state-of-the-art performance with minimal overhead.}

Beyond single-teacher settings, several extensions further improve distillation. For example, multi-level logit knowledge distillation (MLKD)~\cite{MLKD} aligns predictions at multiple levels, while multi-teacher KD with reinforcement learning (MTKD-RL)~\cite{MTKD_RL} dynamically weights multiple teachers to overcome suboptimal fixed weighting.
\textcolor{blue}{Li \textit{et al.}~\cite{li2024dual} proposed DTSKD, a dual-teacher self-KD framework leveraging lightweight branches and MFM to mitigate semantic gaps with high efficiency.}
\textcolor{blue}{Beyond image classification, KD has also been extended to time series analysis. For instance, Boileau \textit{et al.}~\cite{boileau2025interpretabletimeseriesfoundation} distilled reasoning knowledge from foundation models into lightweight language models to enhance temporal interpretability, while Dehigahawattage \textit{et al.}~\cite{Hewa_Dehigahawattage_2025} proposed TSD to extract temporal saliency and further improve fidelity and interpretability.}

In contrast to the aforementioned methods, \algname introduces a feedback mechanism that adaptively refines the teacher model based on feedback from the student. This dynamic adjustment enables the teacher to generate knowledge better aligned with the student’s learning capacity, effectively narrowing the teacher-student capacity gap.

\section{Adaptive Reciprocal Knowledge Distillation}

In this section, we first introduce the motivation and preliminaries underlying our proposed \algname, including the limitations of conventional knowledge distillation under large teacher–student capacity gaps. We then describe our three-stage framework, consisting of initial learning, reciprocal learning, and final learning, and formulate the corresponding optimization objectives. Finally, we provide an analysis of why reciprocal learning can alleviate category correlation collapse and improve the compatibility of the transferred knowledge with the student model.

\begin{figure*}[tb!]
    \centering
    \includegraphics[scale=0.52  ]{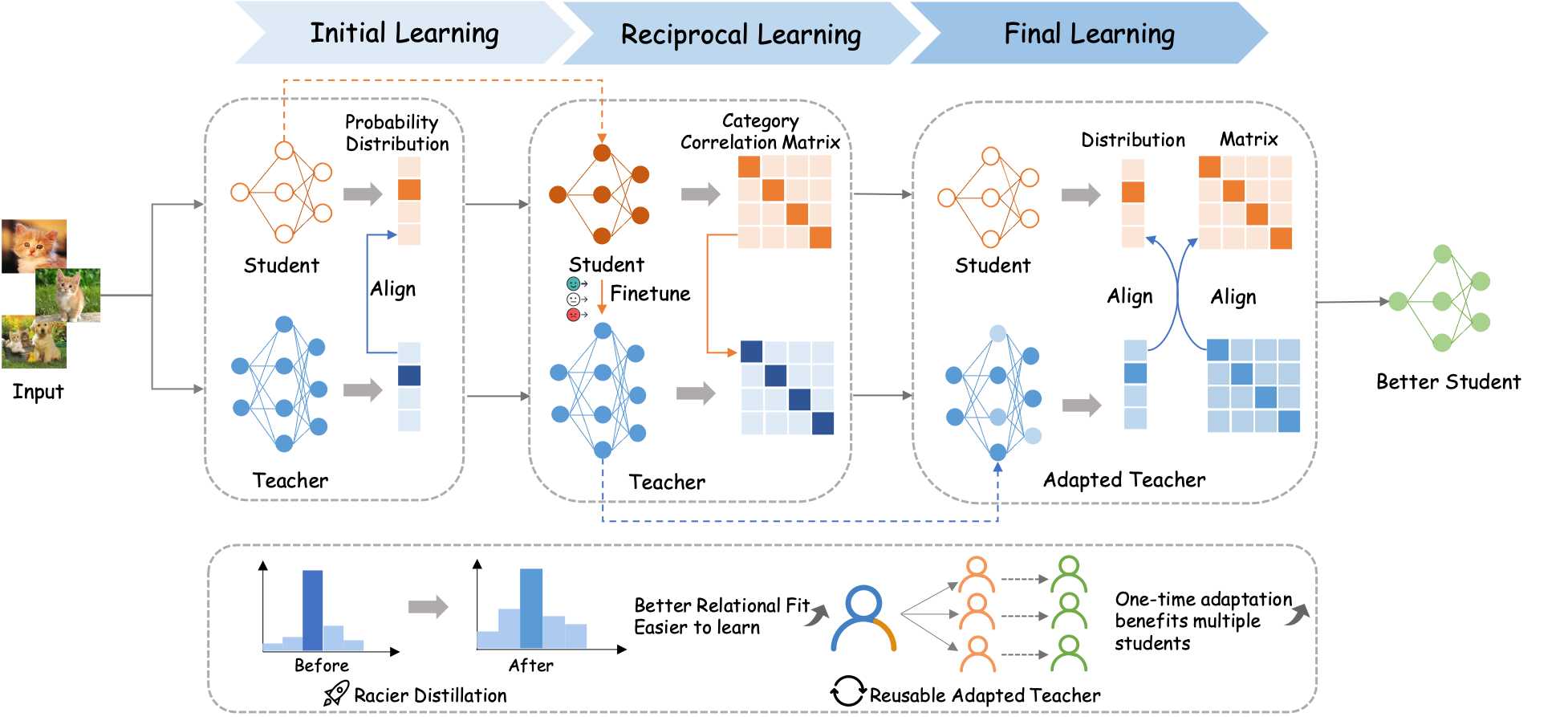}
    \caption{\algname performs reciprocal distillation in three stages: (left) traditional KD, (middle) category correlations based teacher adaptation guided by a frozen student, and (right) re-distillation with the adapted teacher. This bidirectional process generates simplified and compatible knowledge tailored to the student’s learning capacity. }\label{fig:figure2}
\end{figure*}
\subsection{Motivation and Notations}
We start from the classical KD method proposed in \cite{KD_Hinton}. To illustrate the procedure of KD, we consider a $C$-way classification task and denote the logit output of a single input as $z=[z_1,z_2,...,z_C] \in \mathbb{R}^{1\times C}$, then the probability of the $i$-th class can be calculated by
\begin{equation}
    p_i = \frac{e^{z_i / \tau}}{\sum_{j=1}^{C} e^{z_j / \tau}},
\end{equation}
where $z_i$ represents the logit of the $i$-th class and $\tau$ is the temperature scaling hyperparameter.

According to previous work, the loss function of classical KD can be reformulated by
\begin{equation}\label{DKD}
\begin{aligned}
        L_{KD} &=\text{KL}(p^{tea}||p^{stu})    \\
        &= p_t^{tea} \log\left( \frac{p_t^{tea}}{p_t^{stu}} \right) + \sum_{i=1, i \neq t}^{C} p_i^{tea} \log\left( \frac{p_i^{tea}}{p_i^{stu}} \right) \\
        &=\text{KL}(\mathbf{b}^{tea} \| \mathbf{b}^{stu}) + (1 - p_t^{tea}) \text{KL}(\hat{p}^{tea} \| \hat{p}^{stu}),
\end{aligned}
\end{equation}

where ${tea}$ and ${stu}$ denote the teacher and student models, respectively. $p^{{stu}}$ and $p^{{tea}}$ are the predicted distributions of the student model  and the teacher model. We define the information content of the teacher model by its entropy as
\begin{equation}
    H(\mathcal{T}) = -\sum_{i=1}^{C} p^{tea}_i \log p^{tea}_i. 
\end{equation}
\indent KD can be viewed as an information transfer process from the teacher model to the student model, quantified by the mutual information
\begin{equation}
    I(\mathcal{T}; \mathcal{S}) = H(\mathcal{T}) - H(\mathcal{T} \mid \mathcal{S}).
\end{equation}
\indent As can be observed, when $\max_i p^{tea}_i \rightarrow 1$, the coefficient associated with non-target class relational terms in \eqref{DKD} tends to zero and $H(\mathcal{T}) \rightarrow 0$, indicating a substantial reduction in the information content encapsulated within the teacher model's outputs; consequently, the achievable mutual information $I(\mathcal{T}; \mathcal{S})$ is inherently limited, which renders it extremely challenging for the student model to approximate the performance of the teacher model.

\subsection{Overview of \algname}

\subsubsection{Problem Formulation}
\figurename~\ref{fig:figure2} illustrates an overview of our proposed \algname, which includes three stages: initial learning, reciprocal learning, and final learning. The details of these stages are summarized below:
First, we explicitly model the dark knowledge, namely the category correlation inherent in the classification task. We state that model predictions can depict the relationships between classes. In a single prediction, if a class is similar to the ground truth, its corresponding prediction value will be relatively higher than those of other classes, indicating that similar classes exhibit high co-activation in the predictive distribution. Such class correlations can be modeled through predictions on the data as
\begin{equation}\label{G_define}
    G = p^\top p, \, G_{ab} = \sum^B_{k=1}{p_{k,a} \cdot p_{k,b}},
\end{equation}
where $G$ is a $C \times C$ matrix, $B$ is the batch size, $p$ indicates the predictions of the model, and $G_{ab}$ represents the probability that the inputs in this batch are classified into the $a$-th category  and the $b$-th category simultaneously, which quantifies the relationship between the two classes.
\\
\indent However, when the teacher model exhibits excessively high prediction confidence on some  samples, it leads to an extremely sharp output distribution, i.e., the vast majority of the probability mass concentrates on the target class, while the probabilities of non-target classes approach zero. Under such circumstances, the values of the off-diagonal elements in the class correlation matrix computed by \eqref{G_define} become extremely small, due to the minimal degree of simultaneous activation across different classes. This causes the class correlation matrix to degenerate into an approximately diagonal matrix, thereby losing the rich semantic relational information among classes. 
\subsubsection{Stage 1: Initial Learning}
In the initial stage, we employ a traditional KD approach to train a preliminary student model, which acts as a baseline representation to be refined in subsequent stages. The overall loss function for this stage is formulated as 
\begin{equation}
\begin{aligned}
    L_{Init} &= (1 - \alpha) \cdot L_{\text{CE}} + \alpha \cdot L_{KD}\\
    &=(\alpha -1) \cdot \sum_{i=1}^C y_i \log p_i^{{stu}} +\alpha \cdot\sum_{i=1}^C p_i^{tea} \log\left( \frac{p_i^{tea}}{p_i^{stu}} \right),
\end{aligned}
\end{equation}
where $y_i$ denotes the ground‑truth probability of the $i$-th class, $p_i^{stu}$ is the predicted probability of the $i$-th class by the student model, $p_i^{tea}$ is the predicted probability of the $i$-th class by the teacher model, and $\alpha$ is the weighting coefficient that balances the two loss terms.
\subsubsection{Stage 2: Reciprocal Learning} 

To alleviate the issues discussed in the Problem Formulation section, we propose Reciprocal Learning to enhance the information content of the class correlation matrix by adaptively smoothing the predictive distribution of the teacher model. Specifically, we construct a feedback loss function that adapts and optimizes the teacher model under the premise of a fixed student model. This approach reduces the excessive output confidence of the teacher model and enhances the information entropy of category relationships in its outputs, thereby providing the student model with more abundant knowledge.

\begin{equation}
L_{\text{RL}} = \underbrace{(1-\alpha)\cdot L_{\text{CE}}}_{\text{Preservation}} + \underbrace{\alpha\cdot \frac{1}{C}\left\| G^{\text{tea}} - G^{\text{stu}} \right\|_F^2}_{\mathclap{\text{Adaptation}}}\text{,}
\end{equation}
where $G^{{tea}}$ and $G^{{stu}}$are the category correlation matrix computed from the teacher's and student's predictions, respectively.

\subsubsection{Stage 3: Final Learning}
In this stage, the prediction distribution of the new teacher model has been smoothed, and the class correlation matrix contains abundant inter-class knowledge, which better matches the current learning capacity of the student model. Meanwhile, we align the student model with both the KL divergence of the new teacher model and the class correlation matrix of the teacher model, thereby helping the student model absorb knowledge more efficiently. The final distillation loss function is defined as 
\begin{equation}
\begin{aligned}
    L_{Final}
    &=L_{KD}+\frac{1}{C} \left\| G^{{tea}} - G^{{stu}} \right\|_F^2,
\end{aligned}
\end{equation}
where $L_{KD}$ denotes the KL divergence. The overall loss function in Stage 3 is defined as 
\begin{equation}
\begin{aligned}
    L_{ARKD}=
    (1 - \alpha) \cdot L_{\text{CE}} + \alpha \cdot L_{Final},
\end{aligned}
\end{equation}
where $\alpha$ is the weighting coefficient that balances the two loss terms.
\section{Why Does Reciprocal Learning Work?}
In this section, we provide an intuitive explanation of why the proposed reciprocal learning mechanism can improve knowledge transfer. Rather than presenting rigorous theoretical guarantees, we focus on identifying key properties of \algname from information-theoretic and optimization perspectives. These properties help us understand when and why adapting the teacher via category correlation structure alignment is beneficial.

\subsection{Preliminaries and Assumptions}
Consider a $C$-way classification task. Let $p^{tea}, p^{stu} \in \mathbb{R}^{1\times C}$ denote the probability distributions of the teacher and student models for a given input, respectively. Following standard KD, we view distillation as a process of transferring information from teacher model $\mathcal{T}$ to student model $\mathcal{S}$ .
We adopt the following assumptions:

\textbf{Teacher Overconfidence.}
A well-trained large teacher often produces sharply peaked distributions, i.e.,
\begin{equation}
    \max_i p^{tea}_i \to 1, \quad \left\|p^{tea}\right\|_2^2 \to 1,
\end{equation}
which implies low entropy $H(\mathcal{T})$.

\textbf{Student as Noisy Approximation.}
The student prediction can be modeled as a noisy transformation of the teacher signal~\cite{diffkd}
\begin{equation}
    p^{stu} = f(p^{tea}) + \varepsilon,
\end{equation}
where $\varepsilon$ is a zero-mean noise term reflecting capacity and optimization limitations.

Under these assumptions, the effectiveness of distillation depends on how much information about $\mathcal{T}$ is preserved in $\mathcal{S}$, quantified by the mutual information
\begin{equation}
    I(\mathcal{T};\mathcal{S}) = H(\mathcal{T}) - H(\mathcal{T}|\mathcal{S}).
\end{equation}

\subsection{Category Correlation Collapse Under Overconfident Teacher}

According to Section 3.2, for convenience of derivation, we analyze a single input of a batch, so we have the class correlation matrix for an instance as
\begin{equation}
    G = p^\top p, \quad G_{ab} = p_a p_b.
\end{equation}

\textbf{Relational Degeneracy.}
If the teacher distribution is sharply peaked, then the off-diagonal mass of $G^{tea}$ collapses 
\begin{equation}
    \sum_{a \neq b} G^{tea}_{ab} = 1 - \left\|p^{tea}\right\|_2^2 \to 0.
\end{equation}

\textbf{Proof.}
Since $\sum_i p_i = 1$,
\begin{equation}
    \sum_{a \neq b} p_a p_b = \left( \sum_i p_i \right)^2 - \sum_i p_i^2 = 1 - \left\|p\right\|_2^2. 
\end{equation}

When $\max_i p_i \to 1$, we have $\left\|p\right\|_2^2 \to 1$, and thus the off-diagonal sum vanishes.

This implies that highly confident teachers encode little category correlation structure. The class correlation matrix degenerates into an approximately diagonal form, removing informative dark knowledge about class similarities.

\subsection{Reciprocal Learning as Entropy and Compatibility Regularization}

In Stage 2, \algname adapts the teacher by minimizing
\begin{equation}
    L_{RL} = (1-\alpha) \cdot L_{\text{CE}}+\alpha \cdot \left\|G^{tea} - G^{stu}\right\|_F^2,
\end{equation}
with $p^{stu}$ fixed.

Expanding the relational term gives
\begin{equation}
    \left\|G^{tea} - G^{stu}\right\|_F^2
= \left\|p^{tea}\right\|_4^4 + \left\|p^{stu}\right\|_4^4 - 2 (p^{tea \top} p^{stu})^2.
\end{equation}

\textbf{Overconfidence Suppression.}
Minimizing \newline $\left\|G^{tea} - G^{stu}\right\|_F^2$ with respect to $p^{tea}$ penalizes large $\left\|p^{tea}\right\|_2$ and thus discourages overly concentrated teacher distributions.

\textbf{Proof.}
Since $\left\|p\right\|_4^4 \ge \frac{1}{C} \left\|p\right\|_2^4$ and $\left\|p\right\|_2^2 \to 1$ for peaked distributions, the objective increases rapidly as $p^{tea}$ becomes concentrated. Therefore, the optimization favors smoother distributions with higher entropy. 
The adapted teacher $\mathcal{T}'$ obtained after reciprocal learning satisfies
\begin{equation}
    H(\mathcal{T}') > H(\mathcal{T}).
\end{equation}

 \begin{figure}[tb!]
    \centering
    \includegraphics[scale=0.34]{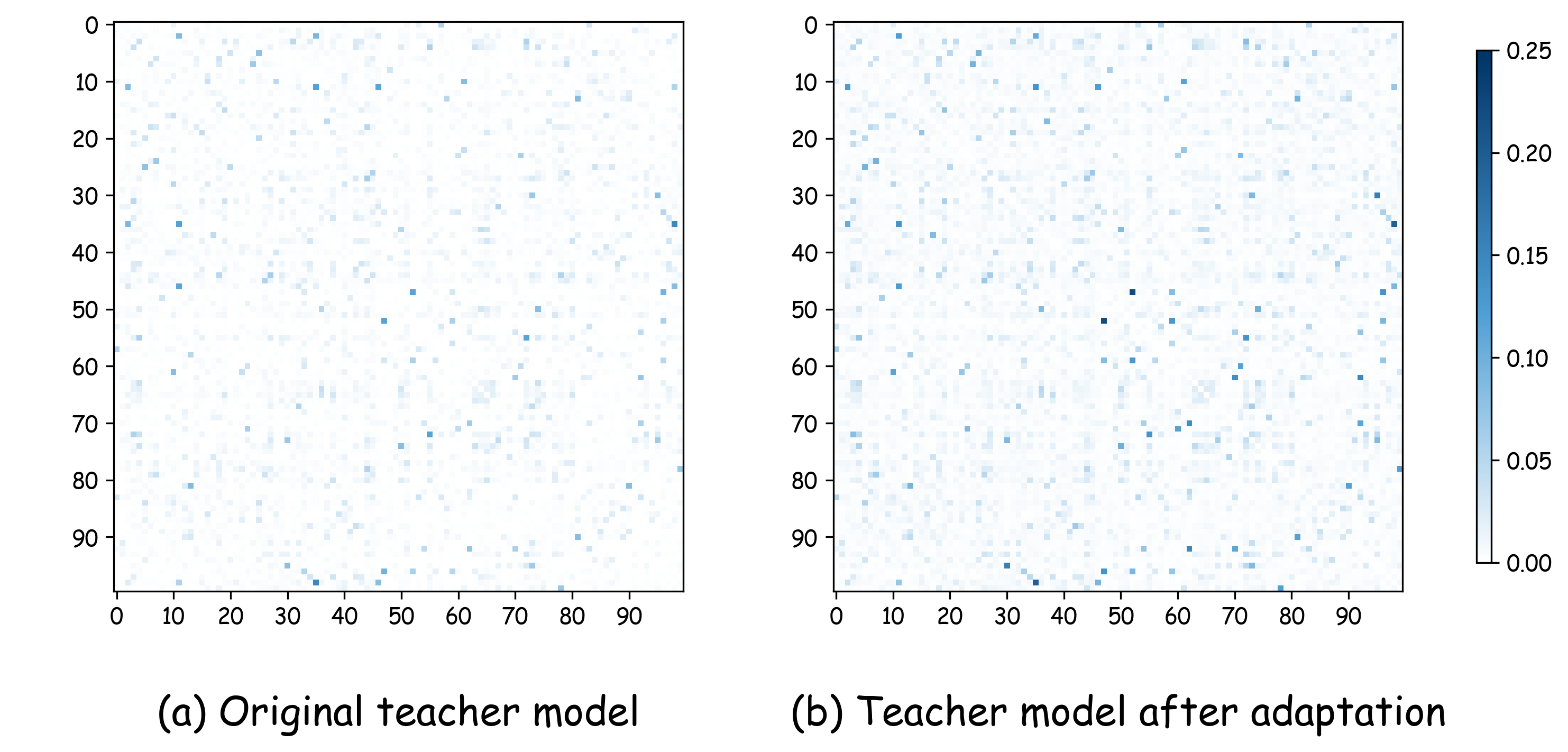}
    \caption{Class correlation matrix heatmaps for a single sample.  
}
    \label{fig:heatmap}
\end{figure}

Thus, reciprocal learning explicitly increases the entropy of teacher outputs while preserving task correctness via the CE term.

As shown in \figurename~\ref{fig:heatmap}, the color intensity represents the strength of category correlation. The off-diagonal regions of the adapted teacher $\mathcal{T}'$ are no longer pure white, indicating that $\mathcal{T}'$ retains richer category correlation knowledge and thus contains more information compared to the original teacher $\mathcal{T}$.

\subsection{Implications for Mutual Information and Student Learning}

\textbf{Mutual Information Enlargement.}
Under Assumption 2, increasing $H(\mathcal{T})$ while maintaining bounded $H(T|S)$ leads to larger mutual information $I(\mathcal{T};\mathcal{S})$
\begin{equation}
    I(\mathcal{T}';\mathcal{S}) = H(\mathcal{T}') - H(\mathcal{T}'|\mathcal{S}) \ge I(\mathcal{T};\mathcal{S}).
\end{equation}

\textbf{Proof.}
Since $H(\mathcal{T}'|\mathcal{S})$ is dominated by the student noise term $\varepsilon$, and reciprocal learning does not increase $\varepsilon$, raising $H(\mathcal{T}')$ increases $I(\mathcal{T}';\mathcal{S})$ directly.

\section{Experiments}

In this section, we conduct extensive experiments to evaluate the effectiveness of \algname. We first introduce the datasets and implementation settings, followed by comparisons with state-of-the-art knowledge distillation methods. We then conduct ablation studies and additional analysis to examine the contributions of different components, the effectiveness of reciprocal teacher refinement, and the compatibility of \algname with feature-based distillation methods. Finally, we visualize the learned representations to provide further insights into the effectiveness of the proposed method.

\subsection{Datasets and Settings}
We evaluate \algname on two widely used image classification benchmarks: CIFAR-100~\cite{CIFAR100} and ImageNet-1k~\cite{ImageNet}.
CIFAR-100 consists of 32×32 RGB images from 100 object categories, with 50,000 images for training and 10,000 images for testing, and is commonly adopted for benchmarking methods on small-scale image classification tasks.
ImageNet-1k is a large-scale image classification dataset comprising approximately 1.2 million training images and 50,000 validation images across 1,000 object categories.
Following standard practice, all ImageNet images are resized to $224\times224$ during training, enabling evaluation under a high-resolution and large-category setting. 

\subsection{Implementation Details}
For CIFAR-100, we evaluate \algname under two settings: (1) Homogeneous architecture, where the teacher and student share the same architecture type (e.g., VGG13 $\to$ VGG8), and (2) Heterogeneous architecture, where the two differ structurally (e.g., VGG13 $\to$ MobileNetV2). Our experiments cover a range of classical models, including ResNet, WRN, VGG, ShuffleNet-V1/V2, and MobileNetV2\cite{resnet_he,WRN,VGG,shufflenet,mobileNet}.

For ImageNet-1k, we consider two representative teacher--student configurations: Res34$\to$Res18 for the homogeneous architecture setting and Res50$\to$MobileNetV2 for the heterogeneous setting. All experiments are conducted under the same hardware environment and follow the standard ImageNet-1k training protocol. Performance is reported in terms of Top-1 and Top-5 accuracy, in accordance with the conventional evaluation protocol for image classification.\newline
\indent We use the SGD optimizer for 240 training epochs, with an initial learning rate of 0.01 for CIFAR-100 and 0.1 for ImageNet. Specifically, for the second stage, we adopt an initial learning rate of 0.001 and perform fine-tuning training for 40 epochs. The distillation temperature $\tau$ is set to 4. All training is conducted on A100 GPUs. Each experiment is run five times with five different random seeds to eliminate randomness, and we report the averaged results for reliability. T-tests are conducted on all experimental results, and all performance improvements achieved by our AR-KD framework are statistically significant with $p$-value < 0.05.

\definecolor{rank1}{RGB}{247,150,89}    
\definecolor{rank2}{RGB}{97,138,210}    
\definecolor{rank3}{RGB}{169,209,142}    

\begin{table*}[tb!]
\centering
\caption{Comparison of Top-1 accuracy (\%) with different methods on CIFAR-100. Note: R32x4, R8x4, R56, R50, R20, W40-2, W40-1, W16-2, MV2, SV1, and SV2 stand for ResNet32x4, ResNet8x4, ResNet56, ResNet50, ResNet20, WRN-40-2, WRN-40-1, WRN-16-2, MobileNetV2, ShuffleNetV1, and ShuffleNetV2.}
\label{tab:CIFAR100}
\scalebox{0.96}{
\renewcommand{\arraystretch}{1}
\setlength{\tabcolsep}{6pt} 

\begin{tabular}{ll|ccccc|ccccc|c}
\hline
\noalign{\vskip 2pt} 
                     &          & \multicolumn{5}{c|}{Homogeneous architecture} & \multicolumn{5}{c|}{Heterogeneous architecture} \\ 
\noalign{\vskip 2pt} 
\hline
\noalign{\vskip 2pt}
\multirow{4}{*}{Method}  &Teacher  & R32x4   & R56     & W40-2   & W40-2  & VGG13  & R32x4   & W40-2   & VGG13   & R50     & R32x4 &\multirow{4}{*}{Avg.} \\
                         &          & 79.42   & 72.34   & 75.61   & 75.61  & 74.64  & 79.42   & 75.61   & 74.64   & 79.34   & 79.42  \\
                         &Student  & R8x4    & R20     & W40-1   & W16-2  & VGG8   & SV1     & SV1     & MV2     & MV2     & SV2    \\
                         &          & 72.50   & 69.06   & 71.98   & 73.26  & 70.36  & 70.50   & 70.50   & 64.60   & 64.60   & 71.80  \\ 
\noalign{\vskip 2pt}
\hline
\noalign{\vskip 2pt}
\multirow{6}{*}{Feature} & FitNet (ICLR15')   & 73.50   & 69.21   & 72.24   & 73.58  & 71.02  & 73.59   & 73.73   & 64.14   & 63.16   & 73.54  &70.77\\
                         & CRD (ICLR20')      & 75.51   & 71.16   & 74.14   & 75.48  & 73.94  & 75.11   & 76.05   & 69.73   & 69.11   & 75.65  &73.59\\
                         & WCoRD (CVPR21')    & 75.95   & 71.56   & 74.73   & 75.88  & 74.55  & 75.40   & 76.32   & 69.47   & 70.45   & 75.96  &74.03\\
                         & ReviewKD (CVPR22') & 75.63   & 71.89   & 75.09   & 76.12  & 74.84  & 77.45   & 77.14   & 70.37   & 69.86   & 77.78  &74.62\\
                         & NORM (ICLR23')     & 76.49   & 71.35   & 74.82   & 75.65  & 73.95  & 77.42   & 77.06   & 68.94   & 70.56   & 78.07  &74.43\\
\noalign{\vskip 2pt}
\hline
\noalign{\vskip 2pt}
\multirow{8}{*}{Logit}   &KD (NIPS14')      & 73.33   & 70.66   & 73.54   & 74.92  & 72.98  & 74.07   & 74.83   & 67.37   & 67.35   & 74.45  &72.35\\
                         &TAKD (AAAI20')     & 73.81   & 70.83   & 73.78   & 75.12  & 73.23  & 74.53   & 75.34   & 67.91   & 68.02   & 74.82  &72.74\\ 
                         & DKD (CVPR22')      & 76.32   & 71.97   & 74.81   & 76.24  & 74.68  & 76.45   & 76.70   & 69.71   & 70.35   & 77.07  &74.43\\
                         & CTKD (AAAI23')     & 73.70   & 71.19   & 73.93   & 75.45  & 73.52  & 74.48   & 75.78   & 68.46   & 68.47   & 75.31  &73.03\\

                         & LSKD (CVPR24')     & 76.62   & 71.43   & 74.37   & 76.11  & 74.36  & -       & -       & 68.61   & 69.02   & 75.56  & -\\
                         & RCKA (IJCAI24')&76.11&-&75.34&76.51&74.97&76.97&77.21&70.12&-&-&-\\
                         & TeKAP (ICLR25')    & 74.79   & 71.32   & 73.80   & 75.21  & 74.00  & 74.92   & 76.75   & 67.39   & 69.00   & 75.43  &73.26\\
                        
                         &RLD (ICCV25')&76.64&72.00&74.88&-&74.93&-&-&69.97&\textbf{70.76}&77.56&-\\
                          &\textbf{\algname (Ours)}    & \textbf{77.29} & \textbf{72.08} & \textbf{75.51} & \textbf{76.78} & \textbf{75.22}    & \textbf{77.63}     & \textbf{77.52}     & \textbf{70.12}     & 70.02     & \textbf{78.60} & \textbf{75.08} \\ 
\noalign{\vskip 2pt}
\hline 
\noalign{\vskip 2pt}

\noalign{\vskip 2pt}
\multirow{3}{*}{}   
                    &Gap-Base  & 4.79 & 3.02 & 3.53 & 3.52 & 4.86 & 7.13 & 7.02 & 5.52 & 5.4 & 6.8 & 5.16 \\
                    &Gap-\algname &2.13 & 0.26 & 0.10 & -1.17 & -0.58 & 1.79 & -1.91 & 4.52 & 9.32 & 0.82 &1.53\\
\noalign{\vskip 2pt}
                    \hline
                    \rowcolor{green!20}
\noalign{\vskip 2pt}
                    & $\Delta$ (to KD)       &\color{black}{+3.96} & \color{black}{+1.42} & \color{black}{+1.97} & \color{black}{+1.86} & \color{black}{+2.24} & \color{black}{+3.56} & \color{black}{+2.69} & \color{black}{+2.75} & \color{black}{+2.67} & \color{black}{+4.15}  &\color{black}{+2.72}\\
\noalign{\vskip 1pt}
\hline
\end{tabular}
}

\end{table*}

\begin{table}[t]
    \centering
    \renewcommand{\arraystretch}{1}
     \caption{Comparisons of Top-1 and Top-5 accuracy (\%) on ImageNet-1k. The original accuracies of the teacher and student model are also reported.}
    \label{tab:imagenet_perf}
    \resizebox{1\linewidth}{!}{
    \begin{tabular}{cl|cc|cc}
    \toprule
     & & Top-1 & Top-5 & Top-1 & Top-5 \\
    \midrule
    \multirow{4}{*}{Method} &  \multirow{2}{*}{Teacher} & \multicolumn{2}{c|}{ResNet34} & \multicolumn{2}{c}{ResNet50} \\
    & & 73.31 & 91.42 & 76.16 & 92.86 \\
    & \multirow{2}{*}{Student} & \multicolumn{2}{c|}{ResNet18} & \multicolumn{2}{c}{MobileNetV2} \\
    & & 69.75 & 89.07 & 68.87 & 88.76 \\
    \midrule
    \multirow{4}{*}{Feature} 
    & AT (ICLR17')  & 70.69 & 90.01 & 69.56 & 89.33 \\
    & OFD (ICCV19')  & 70.81 & 89.98 & 71.25 & 90.34 \\
    & CRD (ICLR20')  & 71.17 & 90.13 & 71.37 & 90.41 \\
    & ReviewKD (CVPR22')  & 71.61 & 90.51 & 72.56 & 91.00 \\
    & CAT-KD (CVPR23') &71.26 &90.45 &72.24 &91.13 \\
    \midrule
    \multirow{5}{*}{Logit} 
    &KD (NIPS14')  & 70.66 & 89.88 & 68.58 & 88.98 \\
    
    & TAKD (AAAI20') & 70.78 & 90.16 & 70.82 & 90.01 \\
    & DKD (CVPR22')  & 71.70 & 90.41 & 72.05 & 91.05 \\
    & MLKD (CVPR23') & 71.90 & 90.55 & \textbf{73.01} &91.42 \\
    & RLD (ICCV25') &71.29&90.59&72.75&91.18\\
    & \textbf{\algname (Ours)} & \textbf{72.13} & \textbf{90.73} & 72.95 & \textbf{91.43} \\
    \midrule \rowcolor{green!20}
    \multirow{1}{*}{} 
    & $\Delta$ (to KD)  &1.47  &0.85  &4.37  &2.45  \\
    \noalign{\vskip 1pt}

    \bottomrule
    \end{tabular}
    }
   
\end{table}

\subsection{Experimental Result}
\subsubsection{Comparison with State-of-the-art Baselines}
\shen{We validate our proposed \algname against the following state-of-the-art baselines including feature-based distillation methods (FitNet~\cite{feature_FitNet}, AT~\cite{feature_AT}, CRD~\cite{feature_CRD}, WCoRD~\cite{WCoRD}, ReviewKD~\cite{feature_ReviewKD}, NORM~\cite{Norm_iclr}, CAT-KD~\cite{feature_catkd}), and logit-based distillation methods (KD~\cite{KD_Hinton},  TAKD~\cite{TAKD}, DKD~\cite{DKD_2022}, CTKD~\cite{Logit_CTKD}, MLKD~\cite{MLKD}, LSKD~\cite{logit_LSKD}, RCKA~\cite{RCKA}, TeKAP~\cite{tekap}, and RLD~\cite{RLD}). }

Tables~\ref{tab:CIFAR100} and \ref{tab:imagenet_perf} summarize the performance of \algname under both homogeneous and heterogeneous teacher–student settings on CIFAR-100 and ImageNet-1k, respectively. As shown in Table~\ref{tab:CIFAR100}, \algname improves student performance across both homogeneous and heterogeneous settings on CIFAR-100. Compared to baseline students, \algname yields up to 7.13\% absolute gains and outperforms vanilla KD by 1.42\%–4.15\%. \algname achieves comparable or superior performance to feature-based methods while only utilizing output logits, demonstrating the effectiveness and generalizability of our adaptive reciprocal distillation framework.

For ImageNet-1k, in the homogeneous ResNet34 $\to$ResNet18 setup, \algname outperforms MLKD, a strong logit-based method, by 0.23\% in Top-1 accuracy. Meanwhile, in the heterogeneous ResNet50$\to$MobileNetV2 configuration, \algname achieves a Top-1 accuracy that is only 0.06\% lower than MLKD, the best-performing logit-based method, while matching the highest reported Top-5 accuracy. These results demonstrate that \algname remains competitive across different architectural settings on ImageNet-1k, benefiting from a lightweight and architecture-agnostic distillation framework.

\begin{table}[tb!]
  \centering
    \caption{Ablation study result. The experiments are conducted on CIFAR-100, with ResNet32x4 as the teacher model, ResNet8x4 as the student model, and Top-1 accuracy as the evaluation metric.}
  \label{tab:ablation}
  \renewcommand{\arraystretch}{1}
  \resizebox{\linewidth}{!}{
  \begin{tabular}{ccccc}
    \toprule
    Group& Initial Learning & Reciprocal Learning & Final Learning & Acc \\
    \midrule
    1& \checkmark & \ding{55} & \ding{55}  & 73.33 \\
    2& \checkmark  & \ding{55} & \checkmark &  76.24\\
    3& \checkmark & \checkmark & \ding{55}  & 76.15 \\
    4& \checkmark & \checkmark & \checkmark  & \textbf{77.29} \\
    \bottomrule
  \end{tabular}
  }

\end{table}

\begin{table*}[tb!]
    \centering
    \caption{Performance comparison of AR-KD (KL vs. Ours) on CIFAR-100. }~\label{tab:KL}
    \renewcommand{\arraystretch}{1}
    \scalebox{1}{
    \begin{tabular}{c|ccccc|ccccc}
        \toprule
        Teacher Model       & Res32x4 & Res56 & W40-2 & W40-2 & VGG13 & Res32x4 & Res32x4 & W40-2 & VGG13 & Res50 \\
        Accuracy    & 79.42   & 72.34 & 75.61 & 75.61 & 74.64 & 79.42   & 79.42         & 75.61 & 74.64 & 79.34 \\
        Student Model       & R8x4    & R20   & W40-1 & W16-2 & VGG8  & SV1     & SV2           & SV1   & MV2   & MV2   \\
        Accuracy    & 72.50   & 69.06 & 71.98 & 73.26 & 70.36 & 70.50   & 71.82         & 70.50 & 64.60 & 64.60 \\
        \midrule
        AR-KD (KL)      & 76.36   & 72.07 & 75.27 & 76.43 & 74.36 & 75.95   & 77.35         & 76.77 & 69.59 & 69.78 \\
        AR-KD (Ours)         & 77.29   & 72.08 & 75.51  & 76.78 & 75.22 & 77.63 & 78.60         & 77.52 & 70.12 & 70.02 \\
        \bottomrule
    \end{tabular}
    }
    
\end{table*}

While \algname delivers strong and often leading performance on both CIFAR-100 and ImageNet-1k, its relative improvements are more pronounced in homogeneous teacher–student settings than in heterogeneous ones. This trend can be attributed to the fact that \algname operates without exploiting intermediate representations, which are known to be particularly beneficial when transferring knowledge across architectures with large structural discrepancies. Nevertheless, across both small-scale (CIFAR-100) and large-scale (ImageNet-1k) benchmarks, \algname achieves performance that is comparable to or competitive with representative feature-based methods, while maintaining a simpler and more flexible formulation. These results highlight the robustness and general applicability of \algname across diverse datasets and distillation scenarios.

\subsubsection{Comparison with Teacher Model}
We further examine the performance gap between the student and teacher models in our proposed AR-KD method. As shown in Table~\ref{tab:CIFAR100}, we calculate the accuracy difference between the student and teacher models (denoted as Gap-ARKD), where a negative gap indicates that the student model outperforms the teacher model. It can be observed that, with our carefully designed approach, the performance of the student model (\algname) is highly aligned with that of the teacher model, with an average accuracy gap of only 1.53. More notably, the student model even slightly surpasses the teacher in some experimental scenarios.

We observe that if the accuracy gap between the teacher model and the student model is not extremely large without distillation, our method can boost the student model’s accuracy to a level very close to that of the teacher model. However, if the student model itself has extremely low accuracy and a substantial gap from the teacher model, such as ResNet50 $\to$ MobileNetV2, the student model still exhibits a noticeable accuracy gap from the teacher model even after optimization with our method. This phenomenon stems from the student model's limited capacity; once this capacity bound is reached, further improvement becomes exceedingly challenging.

\subsubsection{Ablation Study}
To investigate the contributions of each stage in our method, i.e., initial learning, reciprocal learning, and final learning, we conduct four groups of controlled experiments. As shown in Table~\ref{tab:ablation}, the experimental results are analyzed as follows: In the first group, where only the first stage is adopted for training, the performance of our method is comparable to that of the standard KD method. In the second group, we directly perform the third-stage training without fine-tuning the teacher model. Compared with the first group, this configuration achieves a significant performance improvement, indicating that the category correlation matrix encodes rich inter-class semantic relationships—information that is overlooked when only KL divergence alignment is employed. In the third group, we conduct the third stage by aligning the updated teacher model and student model solely via KL divergence, without incorporating the category correlation matrix as a supplement. This setup also outperforms the first group remarkably, demonstrating that the teacher model fine-tuning in the second stage plays a pivotal role. In the fourth group, where all three stages are integrated, our method achieves superior performance over all other experimental groups. This confirms the importance of all three stages to the overall performance gain.

\subsubsection{Comparison with KL-Based Teacher Refinement}
In the early stage of our framework's design, we attempted to fine-tune and constrain the teacher model by aligning the KL divergence between the teacher and student models in the second stage. The experimental results are presented in Table~\ref{tab:KL}. The results indicate that the fine-tuning method based on the category correlation matrix achieves superior performance. We speculate that this KL divergence-based alignment approach struggles to capture the deep category correlation information; it may even lead to the teacher model being misguided by the student model. This further demonstrates the superiority of our proposed method.

\begin{table}[t]
\centering
\caption{Training time for each batch.}
\begin{tabular}{lcccc}
\toprule
Method & KD & ReviewKD & CRD & AR-KD (Ours) \\
\midrule
Time & 15 ms & 62 ms & 109 ms & 33 ms\\
\bottomrule
\end{tabular}
\label{tab:time}
\end{table}

\subsubsection{Combining with Feature Knowledge Distillation Methods}
\algname is a logit-based distillation method whose core objective is to obtain a teacher model that is better suited to the student model. Therefore, we can integrate \algname with feature-based distillation approaches. As shown in Table~\ref{tab:feature}, we combine ReviewKD~\cite{feature_kr} with \algname to conduct our experiments. The results verify the satisfactory scalability of \algname, and its combination with other distillation strategies can achieve higher performance.

\subsubsection{Time cost}
\algname requires extra category correlation matrix computation in training. We evaluate computational overhead by comparing training time for each batch of representative KD methods on CIFAR‑100 under the same hardware configuration, with homogeneous architectures for the teacher and student models. As shown in Table~\ref{tab:time}, despite the extra matrix operation, \algname skips intermediate layers computations and achieves lower per‑batch training latency than feature‑based distillation methods including CRD and ReviewKD.

\subsection{Visualizations}
\begin{figure}[tb!]
    \centering
    \includegraphics[scale=0.34]{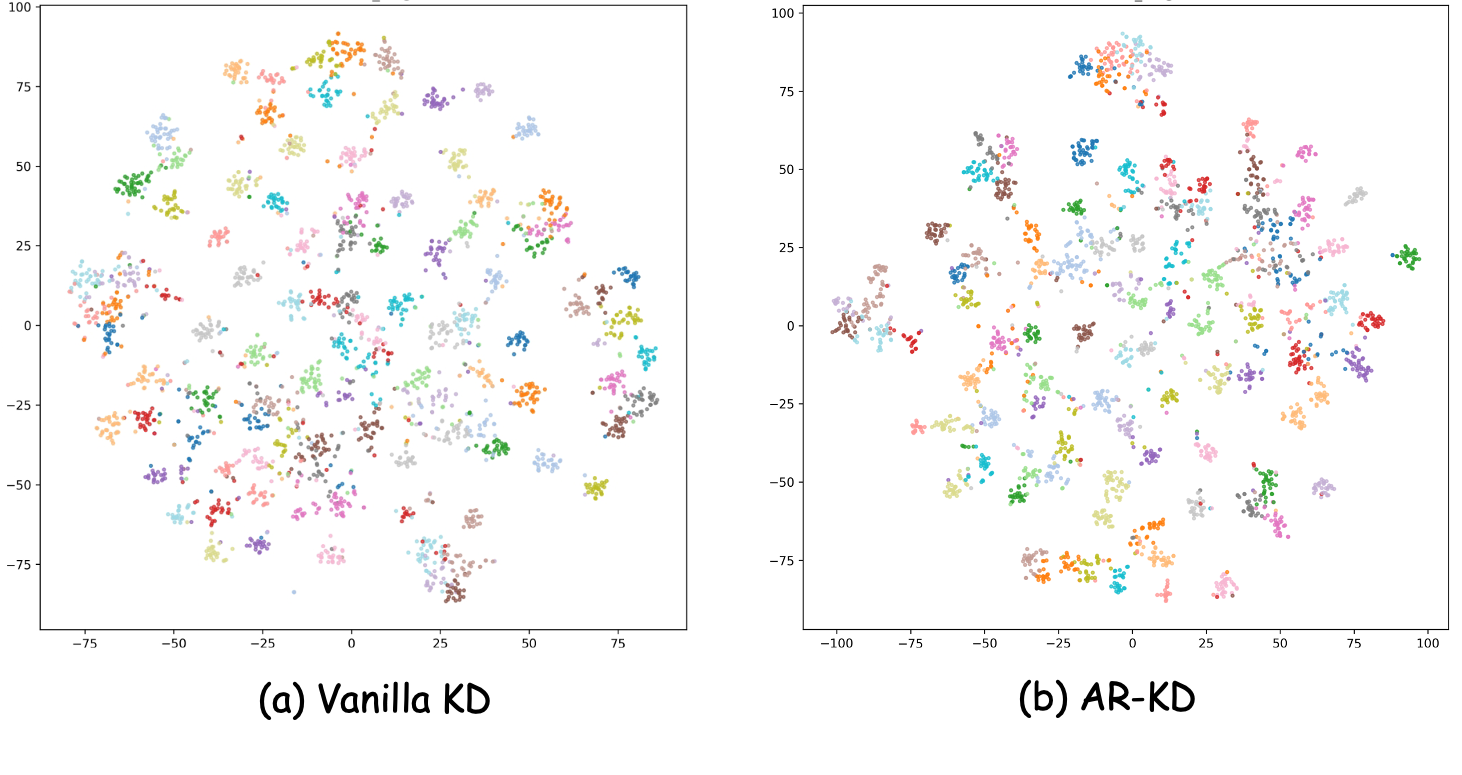}
    \vspace{-0.8cm}
    \caption{Visualization of vanilla KD (left) and \algname (right). }
    \label{fig:figure4}
\end{figure}

 \figurename~\ref{fig:figure4} illustrates the t-SNE visualization of features learned by vanilla KD and \algname on CIFAR-100. It  demonstrates that representations of our method are more separable than vanilla KD, demonstrating that \algname improves the discriminability of deep features.
 
\begin{table}[tb!]
    \centering
    \renewcommand{\arraystretch}{1}
    \caption{Combine with feature KD method on CIFAR-100.}
    \label{tab:feature}
    \scalebox{0.94}{
    \setlength{\tabcolsep}{3pt}
    \begin{tabular}{ccc|ccc}
        \toprule
        Teacher&$\rightarrow$ &Student & ReviewKD & ReviewKD+\textbf{Ours} &$\Delta$\\
        \midrule
        ResNet32x4 &$\rightarrow$ &ResNet8x4 &75.63 &77.65 &\textbf{+2.02}\\
        
        WRN40-2 &$\rightarrow$ &WRN40-1  & 75.09 & 75.88 &\textbf{+0.79}\\
        
        WRN40-2 &$\rightarrow$ &WRN16-2  & 76.12 & 77.08 &\textbf{+0.96}\\

        ResNet56 &$\rightarrow$ &ResNet20    & 71.89 & 72.16 &\textbf{+0.27}\\

        VGG13 &$\rightarrow$ &VGG8   & 74.84 & 75.36 &\textbf{+0.52}\\
        
        \bottomrule
    \end{tabular}
    }
    
\end{table}
\section{Rethinking \algname}

\subsection{Concern About the Scalability of \algname}
A potential concern of \algname is the additional computational overhead introduced by the reciprocal teacher adaptation stage. We clarify that \algname does not retrain the teacher model from scratch. Instead, the second stage only performs lightweight adaptation on a pretrained teacher for a relatively small number of epochs, making the additional optimization cost substantially lower than the original teacher training process.

Moreover, \algname operates purely at the logit level and does not require intermediate feature extraction, feature projection modules, or auxiliary alignment networks commonly used in feature-based distillation methods. Therefore, the overall training pipeline remains simple and computationally efficient.

An important practical advantage of \algname is that the adapted teacher can be reused across multiple student models once reciprocal learning is completed. In practice, we observe that the refined teacher consistently provides more compatible supervision signals for different lightweight students without requiring repeated adaptation. Consequently, the reciprocal learning stage introduces a one-time optimization cost, while its benefits can be amortized throughout the entire distillation pipeline.

From this perspective, \algname achieves a favorable trade-off between computational overhead and distillation effectiveness, particularly under large teacher–student capacity gaps where conventional KD methods often suffer from limited transferability.

 \begin{figure}[tb!]
    \centering
    \includegraphics[scale=0.36]{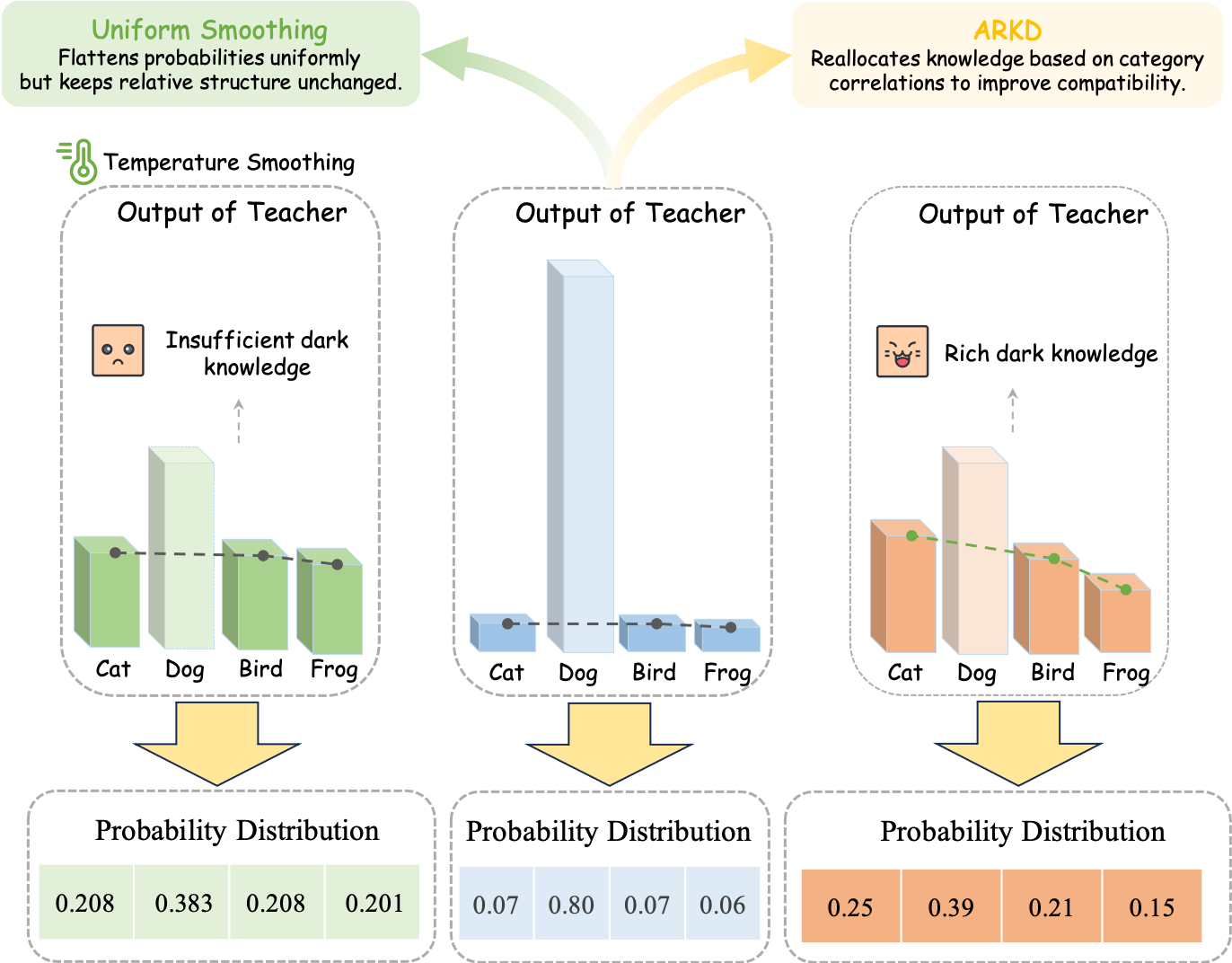}
    \caption{A toy case where two teacher models, $\mathcal{T}_1$ and $\mathcal{T}_2$, adapted in two methods. $\mathcal{T}_1$ merely scales the overall distribution, leaving the interrelationships among non-target classes largely hidden. The original probabilities are 0.07, 0.07, 0.06, and after temperature scaling they remain nearly identical at 0.208, 0.208, 0.201, preserving the same structure. In contrast, $\mathcal{T}_2$ with \algname explicitly restructures the category correlation: frog is actively suppressed to 0.15, creating a clear distinction from cat and bird. This enables the dark knowledge to be effectively transferred.}
    \label{fig:figure5}
\end{figure}

\subsection{Why Simple Temperature Smoothing Is Not Enough}
A natural question arises from our formulation: if overconfident teacher predictions hinder knowledge transfer, why not simply increase the distillation temperature $\tau$ to soften the teacher outputs?

Although temperature scaling can flatten the prediction distribution globally, we argue that it is fundamentally different from the reciprocal adaptation mechanism in AR-KD. Specifically, temperature scaling only rescales the logits uniformly, which changes the sharpness of the probability distribution while preserving the original relational structure among classes. Consequently, the relative similarity relationships encoded by the teacher remain largely unchanged.

In contrast, AR-KD does not merely smooth the output distribution. Instead, it explicitly adapts the category correlation structure of the teacher according to the student representation through correlation matrix alignment. This process modifies how semantic relationships are organized in the teacher outputs, enabling the adapted teacher to generate knowledge that is structurally compatible with the student capacity.

From the perspective of class correlations, temperature scaling mainly enlarges small probabilities uniformly, whereas \algname reshapes the relational dependency among categories. Therefore, the proposed reciprocal learning mechanism can recover richer and more transferable dark knowledge beyond simple entropy smoothing.

\figurename~\ref{fig:figure5} illustrates a toy case comparing two teacher models, $\mathcal{T}_1$ and $\mathcal{T}_2$, adapted with different methods. $\mathcal{T}_1$ merely scales the overall logit distribution via temperature scaling, leaving the interrelationships among non-target classes largely hidden. For instance, the original probabilities for three non-target classes are 0.07, 0.07, 0.06; after scaling they become nearly identical (0.208, 0.208, 0.201), preserving the same flat structure. In contrast, $\mathcal{T}_2$ with \algname explicitly restructures the category correlation: the probability of frog is actively suppressed to 0.15, creating a clear distinction from cat and bird. This restructuring makes the dark knowledge more discernible and thus more effectively transferable to a student model.

\subsection{Rethinking the Capacity Gap in Distillation}

Most existing KD studies typically treat the teacher–student capacity gap primarily as a  performance discrepancy. However, our findings reveal that this gap fundamentally reflects a representation compatibility issue. As the capacity gap increases, the relational structures encoded by the teacher model may exceed the optimization and representation capabilities of the student model, making direct imitation progressively less effective.

From this perspective, the core challenge of KD is not merely ``transferring stronger knowledge", but rather how to adapt complex relational structures into a form that is inherently learnable for the lightweight student models. This  indicates that future KD paradigms should shift their focus from strictly enforcing teacher-student alignment to ensuring representation compatibility. KD achieves its full potential only when the teacher's knowledge is projected into a structural format that the student can successfully assimilate. Ultimately, in the context of extreme capacity gaps, representation compatibility supersedes raw capacity, and targeted adaptation proves more critical than rigid imitation.

\section{Conclusion}
We propose adaptive reciprocal knowledge distillation (\algname), a novel distillation framework that narrows the teacher–student capacity gap by allowing the student to adapt the teacher through reciprocal learning. By aligning category correlation structures via the category correlation matrix, \algname produces smoother and more compatible supervision signals, enabling the student to absorb richer dark knowledge. Extensive experiments on CIFAR-100 and ImageNet-1k demonstrate consistent improvements over state-of-the-art KD baselines, validating the effectiveness and generality of our reciprocal, relationally-aware distillation paradigm.

\section*{GenAI Usage Disclosure}
During the preparation of this work, the authors used AI tools only to improve writing and fix grammar in existing content. After using AI tools, the authors carefully reviewed and edited all changed contents. No AI tools were used for generating research ideas, model configuration choices, or experimental results. The authors take full responsibility for the paper.

\bibliographystyle{ACM-Reference-Format}
\bibliography{sample-base}


\end{document}